\pdfoutput=1

\documentclass[11pt]{article}

\usepackage[final]{acl}

\usepackage{times}
\usepackage{latexsym}

\usepackage[T1]{fontenc}

\usepackage[utf8]{inputenc}

\usepackage{microtype}
\usepackage{inconsolata}

\usepackage{graphicx}
\usepackage{booktabs}
\usepackage{makecell}
\usepackage{multirow}
\usepackage[figuresright]{rotating}
\usepackage[most]{tcolorbox}
\definecolor{BoxBackground}{RGB}{240, 240, 240} 
\definecolor{BoxFrame}{RGB}{0, 0, 0} 
\definecolor{TitleBackground}{RGB}{0, 0, 0} 
\definecolor{TitleText}{RGB}{255, 255, 255} 
\tcbset{
  academicbox/.style={
    boxsep=5pt,
    left=2pt,
    right=2pt,
    bottom=0.5pt,
    boxrule=0.5pt,
    colback=BoxBackground,
    colframe=BoxFrame,
    colbacktitle=TitleBackground,
    coltitle=TitleText,
    enhanced,
    attach boxed title to top left={yshift=-0.1in,xshift=0.1in},
    boxed title style={boxrule=0pt,colframe=white},
    title={#1},
  }
}

\newtcolorbox{AcademicBox}[1][]{academicbox=#1}
\definecolor{SoftBlue}{RGB}{135, 206, 250}  
\definecolor{SoftOrange}{RGB}{255, 224, 178} 
\definecolor{SoftGreen}{RGB}{144, 238, 144}  
\definecolor{CorrectGreen}{RGB}{76, 175, 80} 
\definecolor{ErrorRed}{RGB}{211, 47, 47} 

\title{Rethinking Negative Instances for Generative Named Entity Recognition}

\author{Yuyang Ding$^1$, Juntao Li$^1$\thanks{\; Corresponding author}, Pinzheng Wang$^1$, Zecheng Tang$^1$, Bowen Yan$^2$, Min Zhang$^1$ \\
$^1$Soochow University \; $^2$Tsinghua University \\
\texttt{\{yyding23,pzwang1,zctang\}@stu.suda.edu.cn} \\
\texttt{\{ljt,minzhang\}@suda.edu.cn}, \texttt{yanbw@mail.tsinghua.edu.cn}
}

\begin{document}
\maketitle
\begin{abstract}

Large Language Models (LLMs) have demonstrated impressive capabilities for generalizing in unseen tasks.
In the Named Entity Recognition (NER) task, recent advancements have seen the remarkable improvement of LLMs in a broad range of entity domains via instruction tuning, by adopting entity-centric schema.
In this work, we explore the potential enhancement of the existing methods by incorporating negative instances into training.
Our experiments reveal that negative instances contribute to remarkable improvements by (1) introducing contextual information, and (2) clearly delineating label boundaries.
Furthermore, we introduce an efficient longest common subsequence (LCS) matching algorithm, which is tailored to transform unstructured predictions into structured entities.
By integrating these components, we present GNER, a Generative NER system that shows improved zero-shot performance across unseen entity domains. 
Our comprehensive evaluation illustrates our system's superiority, surpassing state-of-the-art (SoTA) methods by 9 $F_1$ score in zero-shot evaluation.\footnote{Code, datasets, and models are publicly available: \url{https://github.com/yyDing1/GNER}}. 
\end{abstract}

\section{Introduction}

Named Entity Recognition (NER) is a critical and challenging task in the field of Natural Language Processing (NLP).
Previous NER models are constrained by a pre-defined label set and require extensive human annotations, which limits their flexibility and adaptability to unseen entity domains.
Recent advantages in LLMs have enabled the models to be capable of generalizing to unseen tasks~\cite{ouyang2022training,achiam2023gpt} in an auto-regressive generation manner, making it possible to construct powerful NER systems.
However, despite these advancements, recent studies~\cite{wei2023zero,li2023evaluating} show that the zero-shot performance of LLMs still falls behind the supervised training state-of-the-art (SoTA) methods, as LLMs train with limited NER data.

\begin{figure}[!t]
    \centering
    \includegraphics[width=\linewidth]{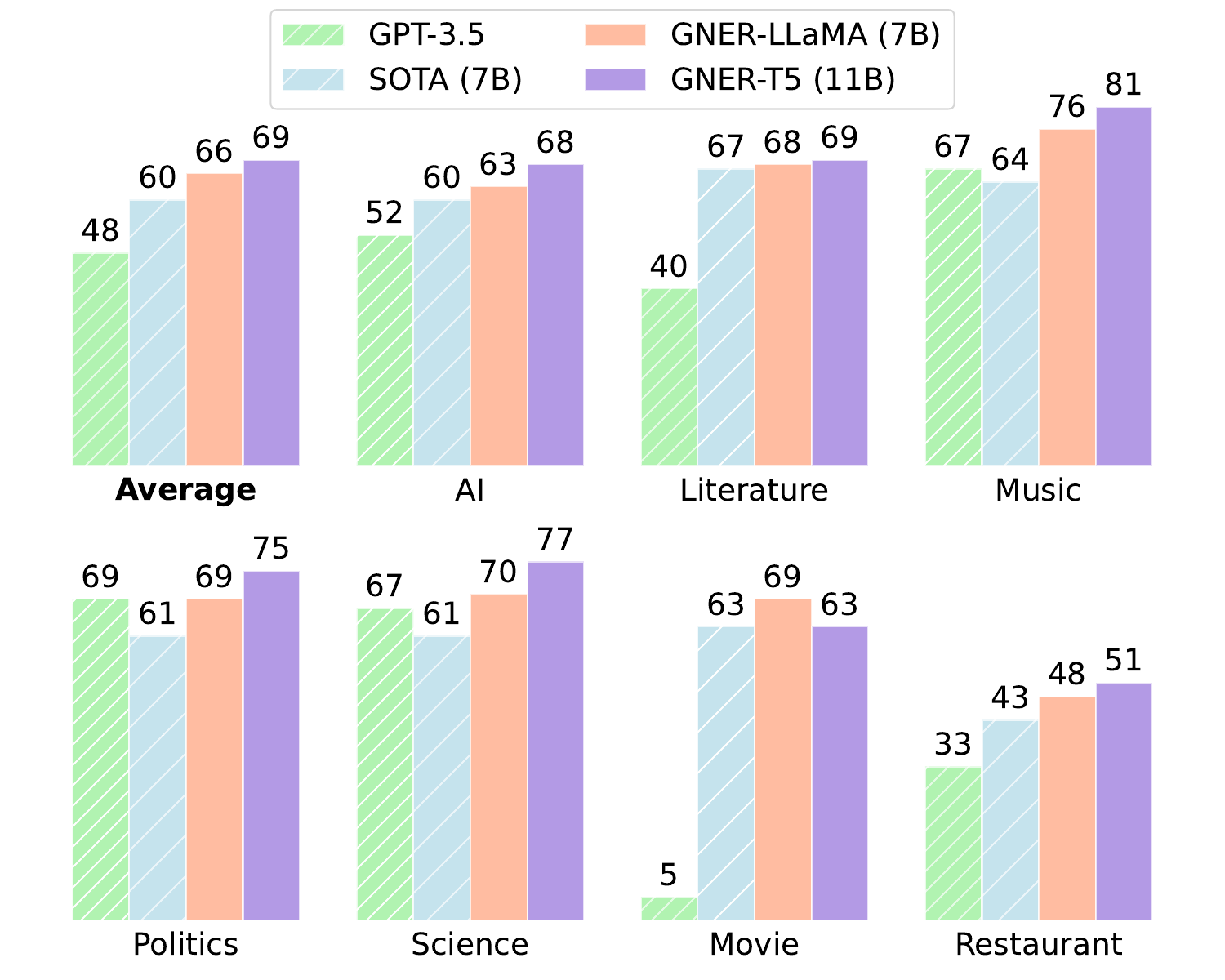}
    \caption{Zero-shot performance of our models. Our models GNER-LLaMA and GNER-T5 both outperform the SoTA~\cite{sainz2023gollie} in zero-shot settings. GPT results are from \citet{zhou2023universalner}.}
    \label{fig:results}
\end{figure}

To bridge this gap, recent works have fine-tuned open-sourced LLMs on diverse NER datasets, enhancing their domain adaptability for NER tasks. 
They utilize varied task schemas to handle NER tasks across multiple domains.
Specifically, InstructUIE~\cite{wang2023instructuie} is fine-tuned on a wide range of IE datasets using a single-round conversation manner.
Meanwhile, UniversalNER~\cite{zhou2023universalner} found that querying all entities at once is less effective than making multiple inquiries, with each inquiry focusing on one entity type at a time.
Additionally, GoLLIE~\cite{sainz2023gollie} enhances zero-shot performance with well-crafted code-style guidelines.
However, these approaches primarily adopt an entity-centric training strategy, focusing on recognizing entities while overlooking the non-entity text, which is crucial as negative instances.
Actually, negative instances play an important role in traditional classification models like BERT Tagging~\cite{devlin2018bert}. 
For generative models, the role of negative instances in the training process has not yet been fully explored.


To calibrate the potential enhancement of including negative instances in training, we first conduct a preliminary study.
Through experiments, we show that negative instances can significantly boost the model's performance by (1) incorporating the contextual information, and (2) enhancing the label boundary between entities and non-entities.
The possible drawback of introducing the negative instances is the increase of the prediction length, leading to inaccurate predictions, reflected by the word addition, omission and substitution.
To tackle the inaccuracy drawbacks, we aim to design a more accurate and efficient algorithm to convert unstructured text into structured entities.

Inspired by the above observations, we design an effective and efficient \textbf{G}enerative \textbf{NER} framework named \texttt{GNER}.
We first design a proper task schema integrating negative instances into the instruction tuning process.
Additionally, we design an LCS Matching algorithm to tackle the issues in the structuring process efficiently.
This innovation ensures accurate categorization and alignment of extracted entities. 
We also demonstrate that zero-shot performance can be enhanced with beam search through a self-correction mechanism.
These strategic developments collectively advance the GNER framework, setting a new standard for accuracy and efficiency in the field of NER.

We conduct experiments on two representative generative models, Flan-T5 and LLaMA.
The resulting models, GNER-T5 and GNER-LLaMA, outperform SoTA by a large margin.
As stated in Fig.~\ref{fig:results}, GNER-LLaMA-7B outperforms the GoLLIE~\cite{sainz2023gollie} trained on Code-LLaMA-7B by 6 $F_1$ score.
Furthermore, compared to the similarly configured model UniversalNER, GNER-LLaMA-7B shows an improvement of 12.7 $F_1$ score, with a 2.5$\times$ boost in inference speed.
We also showcase the potential of smaller models with our 780M GNER-T5-large model, which outperforms all baseline models in both zero-shot and supervised scenarios.

\section{Related Work}

\paragraph{Named Entity Recognition}
Early works format Named Entity Recognition (NER) as a sequence labeling problem~\cite{chiu2016named,huang2015bidirectional,akbik2018contextual,qin2019stack,devlin2018bert}, utilizing the BIO-Tagging scheme.
Then, different methods are proposed to address more complex scenarios, i.e., nested and discontinuous NER.
These methods regard NER as question answering~\cite{li2020unified,mengge2020coarse}, span classification~\cite{fu2021spanner,li2020empirical}, dependency parsing~\cite{yu2020named}, word-level relation classification~\cite{li2022unified}, and so on.
In most of these approaches, negative instances have played a crucial role in the training process, either by integrating all negative instances or employing sampling methods to select part of them~\cite{li2022rethinking}.
However, the performance of the above-mentioned supervised models significantly decreases in zero-shot settings~\cite{liu2021crossner}, especially when the data and domain distribution significantly diverge from those seen of training.

\paragraph{Zero-shot NER}

\begin{figure}[!t]
    \centering
    \includegraphics[width=\linewidth]{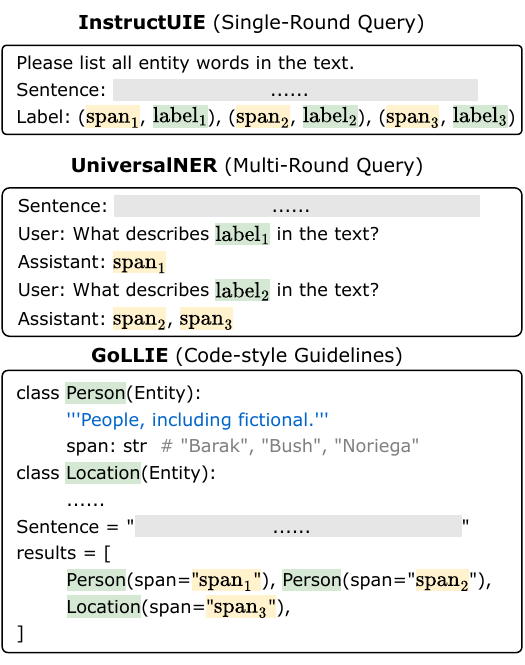}
    \caption{A simplified example of instructions in InstructUIE~\cite{wang2023instructuie}, UniversalNER~\cite{zhou2023universalner} and GoLLIE~\cite{sainz2023gollie}.}
    \label{fig:task_schema}
\end{figure}

Instruction tuning~\cite{wei2021finetuned,chung2022scaling}, also known as multi-task fine-tuning, has emerged as a leading method to achieve generalization to unseen tasks by fine-tuning pre-trained LLMs on a diverse collection of tasks phrased as text-to-text problems~\cite{longpre2023flan}.
In NER, numerous works have explored the potential of LLMs across diverse domains.
For instance, InstructUIE~\cite{wang2023instructuie} is fine-tuned on a wide range of IE datasets and achieves impressive results in both zero-shot and supervised settings.
UniversalNER~\cite{zhou2023universalner} explores the effectiveness of knowledge distillation and multi-round conversational training paradigms in enhancing model generalization, achieving superior results.
GoLLIE~\cite{sainz2023gollie} introduces an innovative strategy by integrating well-crafted code-style guidelines into instructions, which has been found to further improve the model's zero-shot performance.
A simplified version of the task schema of the mentioned methods above is shown in Fig.~\ref{fig:task_schema}.
It can be concluded that the task schema plays an important role in determining the learning paradigm of models, significantly influencing their performance.
We observe that these methods are \textbf{entity-centric}, meaning only the entity portions are involved in training and used for backpropagation to train the model.

\section{Incorporating Negative Instances}
\label{sec:preliminary}

In this section, we start from the entity-centric schema and explore the possible improvements by incorporating negative instances.
Through experiments, we demonstrate the impact of negative instances in mitigating Unlabeled Errors (UE), Noisy Errors (NE), and Boundary Errors (BE) by (1) introducing contextual information and (2) enhancing entity boundaries.


\subsection{Definition \& Settings}
\label{sec:Settings}

The Named Entity Recognition (NER) task can be formally defined as a function mapping of input tokens $X = \{x_1, x_2, \ldots, x_n\}$ and a pre-defined set of entity types $L = \{l_1, l_2, \ldots l_m\}$ to entity labels $Y = \{y_1, y_2, \ldots, y_n\}$.
The positive and negative instances can be formulated as follows:
\begin{equation}
\begin{aligned}
    P &= \left\{(x_i, y_i) \mid i \in \{1, \ldots, n\}, y_i \in L \right\}, \\
    N &= \left\{(x_i, y_i) \mid i \in \{1, \ldots, n\}, y_i = O \right\}, \\
\end{aligned}
\end{equation}
where \textit{O} represents non-entity text.


For the experimental setup, we choose Flan-T5-large, a model with 780M parameters, as our backbone.
Additionally, we sample 10K samples from the Pile-NER~\cite{zhou2023universalner} dataset as our training set and 200 samples for each subtask of CrossNER~\cite{liu2021crossner} validation set to evaluate the model's zero-shot performance.
To conduct a more detailed evaluation, in addition to the $F_1$ score, we introduce the following three metrics to assess model performance:

\paragraph{Unlabeled Error (UE)}
The model fails to recognize the entity and labels it as ``O''.

\paragraph{Noisy Error (NE)}
The model mistakenly label an entity with another incorrect entity tag.

\paragraph{Boundary Error (BE)}
The model correctly predicts the entity type but fails to identify its full extent, either capturing only a portion of the entity or resulting in overlaps.



\begin{figure}[!t]
    \begin{AcademicBox}[\footnotesize An example of the constructed prompt]
    \small
    \textbf{Token inputs ($X$):} John explored Tokyo , sampling its famed sushi , and flew back to New York . \\
    \textbf{Entity type ($L$):} [Person, Location] \\
    \vspace{-5pt} \hrule \vspace{4pt}
    \textbf{Training prompt} \\
    \textbf{w/o context (entity-centric):} \\
    \quad [John](Person) [Tokyo](Location) [New York](Location) \\
    \textbf{w/ context length 1:} \\
    \quad [John](Person) explored [Tokyo](Location) , ...... to [New York](Location) . \\
    \textbf{w/ full context:} \\
    \quad [John](Person) explored [Tokyo](Location) , sampling its famed sushi , and flew back to [New York](Location). \\
    \textbf{w/ full context and label boundary (BIO):} \\
    John(B-Person) explored(O) Tokyo(B-Location) ,(O) sampling(O) its(O) famed(O) sushi(O) ,(O) and(O) flew(O) back(O) to(O) New(B-Location) York(I-Location) .(O) \\
    \textbf{w/ full context and label boundary (BIOES):} \\
    John(S-Person) explored(O) Tokyo(S-Location) ,(O) sampling(O) its(O) famed(O) sushi(O) ,(O) and(O) flew(O) back(O) to(O) New(B-Location) York(I-Location) .(O)
    \end{AcademicBox}
    \caption{Constructed prompts used for training.}
    \label{fig:negative_boundary_example}
\end{figure}

\begin{table}[!t]
    \centering
    \small
    \begin{tabular}{l|ccc|c}
    \toprule
    & UE & NE & BE & $F_1$ \\
    \midrule
    w/o context (entity-centric) & 7.8 & 16.7 & 3.8 & 59.0 \\
    w/ context length 1 & 7.6 & 15.7 & 3.8 & 60.3 \\
    w/ full context & 7.7 & 14.9 & 3.7 & 61.0 \\
    w/ full context \& BIO & 7.5 & 14.4 & 3.3 & \textbf{61.8} \\
    w/ full context \& BIOES & 7.6 & 14.7 & 3.5 & 61.2 \\
    \bottomrule
    \end{tabular}
    \caption{Unlabeled Error (UE), Noisy Error (NE) and Boundary Error (BE) in our preliminary study.}
    \label{table:error_and_f1}
\end{table}

\subsection{Learning with Entity Context}
\label{sec:LearningwithEntityContext}

\begin{figure}[!t]
    \centering
    \includegraphics[width=\linewidth]{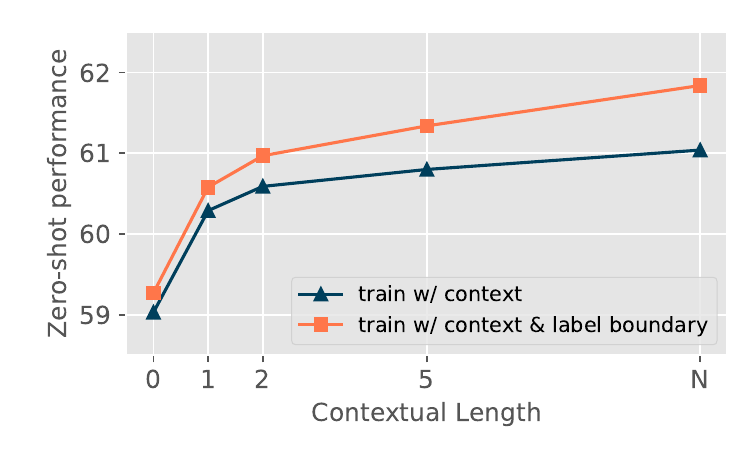}
    \caption{Zero-shot performance of training with entity context and enhanced boundary strategies. A contextual length of 0 indicates no context is included, while a length of $N$ signifies that the entire sentence is included.}
    \label{fig:preliminary_results}
\end{figure}

We integrate the contextual information before and after an entity into our training process to explicitly enable the model to recognize entities based on their surrounding context. 
Specifically, we introduce negative instances that are closest to the entity, extending up to a length $L$, until encountering the boundary of the sentence, as part of our training instances.
An example of our constructed training prompt is shown in Fig.~\ref{fig:negative_boundary_example}, and the corresponding results are summarized in Table~\ref{table:error_and_f1} and Fig.~\ref{fig:preliminary_results}.

The context surrounding entities plays a significant role in determining their categories, with a notable improvement observed when increasing the contextual length from 0 to 1.
As the contextual length increases, performance progressively improves, showing that the model benefits from the context, and the ratio of noisy error (NE) significantly decreases, largely contributing to the improvement in the final $F_1$ score.
Qualitatively, we further analyze through case studies why the closest negative instances contribute to improvement.
We discover that the model tends to learn more from the context, such as ``flew to'' prompting the model to focus more on the following entity ``New York'' instead of merely memorizing ``New York'' as a location. 
We also experiment by placing terms with multiple meanings, such as Jordan, Amazon, and Mercury, after ``flew to'' and observe that the model consistently identified them correctly.




\subsection{Entity Boundary of Generative Model}
\label{sec:EntityBoundaryofGenerativeModel}


The above analysis has demonstrated the effectiveness of entity contexts, where labels are applied exclusively to entity parts.
We then adopt BIO-tagging~\cite{huang2015bidirectional} to enrich all label information, with the training prompt as shown in Fig.~\ref{fig:negative_boundary_example}. 
The results in Table~\ref{table:error_and_f1} indicate that the introduction of BIO tagging effectively strengthens the boundaries around entities, leading to a significant reduction in boundary errors.
As shown in Fig.~\ref{fig:preliminary_results}, compared to training with contexts alone, incorporating the BIO-tagging strategies results in consistent improvements across various contextual lengths.
In addition to BIO tagging, we also try the BIOES tagging scheme, based on the intuition that the BIOES tagging method provides a stronger delineation of entity boundaries.
However, we find that the performance of BIOES tagging is not as good as BIO tagging.
The results are listed in Table~\ref{table:error_and_f1}.
Upon further analysis, we discover that the BIOES tagging seems harder to learn under the auto-regressive generation manner, where each token is predicted sequentially.
For instance, both ``B-'' and ``S-'' can serve as the beginning of an entity, but only ``B-'' can be followed by ``I-'', which may confuse the model for subsequent words.

\subsection{The role of negative instances}

The improvement in model performance ($F_1$ score) can be explained as follows:

\paragraph{Precision and Recall}
To put it more directly, the improvements over entity-centric approaches are primarily reflected in (1) Precision: The context surrounding an entity often leads to a more accurate determination of its type, and (2) Recall: The model is guided to make judgments on every token in a sentence (including those in non-entity texts), which helps recall more entities.
We observe improvements in both recall and precision, which in turn lead to an increase in the $F_1$ score.

\paragraph{Less Unlabeled, Noisy and Boundary Error}
As indicated in Table~\ref{table:error_and_f1}, context helps mitigate Unlabeled and Noisy Errors, while BIO tagging strengthens entity boundaries, thereby reducing Boundary Errors.
These reductions in errors directly contribute to improved performance.


\section{Method}

In this section, we present our GNER framework.
We start by describing our task schema, which integrates negative instances into the training process for better usage of contextual information (section~\ref{sec:LearningwithEntityContext}) and sensitivity to the entity boundaries (section~\ref{sec:EntityBoundaryofGenerativeModel}), followed by the correlated tuning strategies. 
We also propose an effective longest common subsequence (LCS) matching algorithm, to convert the model's unstructured text outputs into structured data efficiently, thereby enhancing the accuracy of our system.

\subsection{Task Schema}

Integrating negative instances, specifically those parts of the sentence labeled as ``O'' to indicate non-entity text, enhances the generative process by including contextual information and the discrimination of entity boundaries, thereby boosting the model's performance, as detailed in section~\ref{sec:preliminary}.
Due to the token-by-token generation paradigm of generative models, we design a token-by-token prediction task schema, where the model predicts the category of each token as it generates them, either entities or non-entities. 
This schema offers a more direct and focused way, where each token is annotated individually and assigned a specific entity label based on its context within the sequence.

\subsection{Instrucion Tuning}

\begin{figure}[!t]
    \begin{AcademicBox}[\footnotesize Instruction Tuning Prompt]
    \small
    \textbf{Task Description:} \\
    Please analyze the sentence provided, identifying the entity type for each word on a token-by-token basis.\\
    Output format is: word\_1(label\_1), word\_2(label\_2), ... \\
    \textbf{Guideline:} \\
    We'll use the BIO-format to label the entities, where: \\
    1. B- (Begin) indicates the start of a named entity. \\
    2. I- (Inside) is used for words within a named entity but are not the first word. \\
    3. O (Outside) denotes words not part of a named entity. \\
    Use the specific entity tags: \fcolorbox{SoftBlue}{SoftBlue}{$l_1, l_2, \ldots$, $l_m$ and O}. \\
    \textbf{Input:} \fcolorbox{SoftOrange}{SoftOrange}{$x_1 \quad x_2 \quad \ldots \quad x_n$} \\
    \textbf{Output:} \fcolorbox{SoftOrange}{SoftOrange}{$x_1(\hat{y}_1) \quad x_2(\hat{y}_2) \quad \ldots \quad x_n(\hat{y}_n)$}
    \end{AcademicBox}
    \caption{Prompt used for instruction tuning.}
    \label{fig:our_data_format}
\end{figure}

\paragraph{Instruction Format}
As shown in Fig.~\ref{fig:our_data_format}, our designed instruction prompt includes four parts: task description, guideline, input, and output.
To enhance our model's ability to generalize across diverse labels and effectively handle real-world data, we implement some regularization strategies: (1) class order shuffling, where the order of entity classes is randomly shuffled, and (2) external entity sampling\footnote{\citet{zhou2023universalner} refers to this as negative entity sampling, which is different from the negative instances discussed in this work. We term it ``external'' to differentiate it.}, involving the entity types that are absent in the given sentence in the training prompt.

\paragraph{Task Adaption \& Supervised Fine-tuning}
Zero-shot capabilities of LLMs in NER are limited due to their exposure to relatively little NER data during training.
To equip the model with capabilities specific to NER tasks, we first perform task adaptation on NER data spanning various domains.
Subsequently, to assess the model's zero-shot capabilities, we evaluate it against unseen entity types.
We proceed to extensively fine-tune our models on a wide range of publicly available NER data, aiming to enhance our model's effectiveness in supervised settings, followed by supervised evaluations.


\subsection{LCS Matching Algorithm}
\label{sec:lcs}
The prediction length of the model increases with the integration of entity contexts and BIO tagging strategies. 
A longer generation sequence might bring challenges to the popular generative LLMs.
The model's output may include omissions, additions, and substitutions of words.
We launch a detailed case study and find the potential causes of these issues:
(1) noise in the original text, (2) missing words in the vocabulary, and (3) accumulative exposure bias.
The representative examples, issue proportion, along with detailed analysis, are documented in Appendix~\ref{sec:longsequencepeoblem}.

To handle these problems, we develop a LCS Matching algorithm that provides a straightforward and effective solution to these challenges.
Formally, given a sentence $X = \{x_1, x_2, \ldots, x_n\}$, the generated outputs can be formatted as $``\tilde{x}_1(\tilde{y}_1)\quad \tilde{x}_2(\tilde{y}_2)\quad \ldots \quad \tilde{x}_m(\tilde{y}_m)"$.
Firstly, we utilize regular expression matching to obtain the predicted sequence $\tilde{X} = \{\tilde{x}_1, \tilde{x}_2, \ldots, \tilde{x}_m\}$ and the corresponding answers $\tilde{Y} = \{\tilde{y}_1, \tilde{y}_2, \ldots, \tilde{y}_m\}$. Due to the inherent uncertainties in generation, $\tilde{X}$ often differs from $X$. 
Next, we establish a one-to-one correspondence between the words in the original sequence $X$ and the generated sequence $\tilde{X}$, then map the labels of the corresponding words $\tilde{Y}$ back to obtain the final prediction results $\hat{Y}$.
A common method involves calculating the Longest Common Subsequence (LCS) between $X$ and $\tilde{X}$ to identify the correspondence between words, using the classic dynamic programming algorithm with time complexity of $O(N^2)$.
Combined with the actual NER task scenarios and our task schema, we make the optimizations in the matching algorithm and condition.
The resulting algorithm can handle these issues effectively with a time complexity of $O(N\log N)$.
The optimization in matching conditions (i.e., Back Tokenization) further enhances the robustness across various models in our system.
More details concerning the optimization can be seen in Appendix~\ref{sec:lcsmatching}.

\section{Experiments}

\begin{table*}[!t]
    \centering
    \footnotesize
    \tabcolsep=1.8mm
    \begin{tabular}{llccccccc|c}
    \toprule
    \bf Model & Backbone & AI & Literature & Music & Politics & Science & Movie & Restaurant & \bf Avg \\
    \midrule
    ChatGPT & - & 52.4 & 39.8 & 66.6 & 68.5 & 67.0 & 5.3 & 32.8 & 47.5 \\
    InstructUIE$^{\dagger}$ & flan-t5-xxl (11B) & 48.4 & 48.8 & 54.4 & 49.9 & 49.4 & \underline{63.0} & 21.0 & 47.8 \\
    GoLLIE-7B$^{\dagger}$ & Code-LLaMA-7B & 60.3 & 67.1 & 64.5 & 60.8 & 60.5 & \underline{63.0} & 43.4 & 59.9 \\
    GoLLIE-13B$^{\dagger}$ & Code-LLaMA-13B & 63.8 & 60.1 & 68.5 & 56.2 & 61.5 & 62.5 & 49.8 & 60.3 \\
    \midrule
    UniNER-7B$^{\ddagger}$ & LLaMA-7B & 53.5 & 59.4 & 65.0 & 60.8 & 61.1 & 42.4 & 31.7 & 53.4 \\
    UniNER-13B$^{\ddagger}$ & LLaMA-13B & 54.2 & 60.9 & 64.5 & 61.4 & 63.5 & 48.7 & 36.2 & 55.6 \\
    GLiNER-L$^{\ddagger}$ & DeBERTa-v3-300M & 60.6 & \underline{68.4} & 69.5 & \underline{74.8} & 69.4 & 57.2 & 42.8 & 63.2 \\
    \midrule
    \multirow{4}{*}{GNER-T5$^{\ddagger}$} & flan-t5-base (250M) & 56.8 & 58.7 & 72.3 & 64.5 & 68.0 & 54.5 & 41.4 & 59.5 \\
    & flan-t5-large (780M) & 62.6 & 58.2 & 76.7 & 67.0 & \underline{72.6} & 58.6 & 48.6 & 63.5 \\
    & flan-t5-xl (3B) & 62.1 & 64.9 & \underline{80.6} & 73.7 & 68.7 & \underline{63.0} & \underline{49.8} & \underline{66.1} \\
    & flan-t5-xxl (11B) & \textbf{68.2} & \textbf{68.7} & \textbf{81.2} & \textbf{75.1} & \textbf{76.7} & 62.5 & \textbf{51.0} & \textbf{69.1} \\
    GNER-LLaMA$^{\ddagger}$ & LLaMA-7B & \underline{63.1} & 68.2 & 75.7 & 69.4 & 69.9 & \textbf{68.6} & 47.5 & \underline{66.1} \\
    \bottomrule
    \end{tabular}
    \caption{Zero-shot evaluation results, where $\dagger$ denotes IE Models and $\ddagger$ denotes NER Models. Results for ChatGPT and UniNER are from \citet{zhou2023universalner}; InstructUIE are from \citet{wang2023instructuie}; GoLLIE are from \citet{sainz2023gollie}; GLiNER-L are from \citet{zaratiana2023gliner}. We bold the best results and underline the second-best results. More details about the performance including error bars are shown in Appendix~\ref{sec:detailed_results}.}
    \label{tab:zero_shot_exp}
\end{table*}


\subsection{Settings}

\paragraph{Datasets}
The datasets used in our experiments include:
(1) \textbf{Task Adaptation Datasets:}
Following the setting of \citet{zhou2023universalner}, we first train our model with Pile-NER, which consists of approximately 240K entities across 13K distinct entity categories.
These passages are sampled from the Pile Corpus~\cite{gao2020pile} and subsequently processed using ChatGPT to generate the inherent entities openly.
To evaluate the model's zero-shot performance in unseen entity types, we adopt two widely-used datasets, i.e., CrossNER~\cite{liu2021crossner} and MIT~\cite{liu2013asgard}.
(2) \textbf{Supervised Datasets:}
Following the task adaptation phase, the performance of the model can be further enhanced by training across a wide range of well-annotated NER datasets~\cite{zhou2023universalner}.
To achieve this, we compile 18 public NER datasets in the BIO format for additional training, subsequently assessing performance on the test splits of these 18 datasets.
From the 20 datasets used in \citet{wang2023instructuie}, we exclude two nested NER datasets, ACE2005 and GENIA, due to their incompatibility with the BIO format. 
Following the settings of \citet{wang2023instructuie}, we randomly select 10K data points from each dataset to create a mixed set. In cases where a dataset contains fewer than 10K samples, we incorporate its entire dataset.
Additional information regarding the datasets is available in Appendix \ref{sec:appendix_data}.

\paragraph{Compared Baselines}
Our main point of comparison is \textbf{UniversalNER}~\cite{zhou2023universalner} as it is the approach closest to our system, with similar data and training procedures.
Another baseline considered for comparison is \textbf{GLiNER}~\cite{zaratiana2023gliner}, which utilizes bi-directional models to match entity types with textual spans in a latent space.
We also include some strong Information Extraction (IE) systems like \textbf{InstructUIE}~\cite{wang2023instructuie}, which is based on Flan-T5-xxl~\cite{chung2022scaling} and fine-tuned on diverse information extraction datasets, and \textbf{GoLLIE}~\cite{sainz2023gollie}, which is based on Code-LLaMA~\cite{roziere2023code}, and use guidelines to improve model's zero-shot performance. We use strict entity-level micro-$F_1$ as the evaluation metric for comparison. 
Previous work lack a uniform setting; UniNER removed the ``else'' entity type in the CrossNER dataset, while InstructUIE, GoLLIE, and GLiNER retained it. 
To standardize the settings, we reevaluated their methods using their publicly released checkpoints and code, ensuring that all test set configurations are consistent with UniNER.

\paragraph{Backbones \& Implementation}
Generative models typically consist of two types of architectures, i.e., the encoder-decoder architecture and the decoder-only architecture. We conduct experiments on both of these architectures. Specifically, we select Flan-T5 (encoder-decoder) and LLaMA (decoder-only) as our backbone models.
To ensure a fair comparison, our training settings for GNER-T5 align with those of InstructUIE~\cite{wang2023instructuie}, and those for GNER-LLaMA are consistent with UniversalNER~\cite{zhou2023universalner}.
Due to our model producing longer output sequences, we implement longer length limits for both input and output.
More details can be found in Appendix \ref{sec:train_settings}.

\begin{table*}[!t]
    \centering
    \footnotesize
    \begin{tabular}{l|c|cc|c|cc|c}
    \toprule
    \bf Method & ChatGPT & InstructUIE & GNER-T5 & \multirow{2}{*}{$\Delta$} & UniNER & GNER-LLaMA & \multirow{2}{*}{$\Delta$} \\
    \bf Backbone & - & \multicolumn{2}{c|}{flan-t5-xxl (11B)} &  & \multicolumn{2}{c|}{LLaMA-7B} &  \\
    \midrule
    AnatEM & 30.7 & 88.52 & \textbf{90.30} & +1.78 & 88.65 & 90.24 & +1.59 \\
    bc2gm & 40.2 & 80.69 & \textbf{84.29} & +3.60 & 82.42 & 83.18 & +0.76 \\
    bc4chemd & 35.5 & 87.62 & \textbf{90.04} & +2.42 & 89.21 & 89.40 & +0.19 \\
    bc5cdr & 52.4 & 89.02 & 89.95 & +0.93 & 89.34 & \textbf{90.27} & +0.93 \\
    Broad Twitter & 61.8 & 80.27 & \textbf{84.56} & +4.29 & 81.25 & 83.74 & +2.49 \\
    CoNLL2003 & 52.5 & 91.53 & 93.28 & +1.75 & 93.30 & \textbf{93.60} & +0.30 \\
    FabNER & 15.3 & 78.38 & 83.20 & +4.82 & 81.87 & \textbf{85.39} & +3.52 \\
    FindVehicle & 10.5 & 87.56 & 97.37 & +9.81 & 98.30 & \textbf{98.62} & +0.32 \\
    HarveyNER & 11.6 & 74.69 & \textbf{76.33} & +1.64 & 74.21 & 74.73 & +0.52 \\
    Movie & 5.3 & 89.58 & 89.28 & -0.30 & 90.17 & \textbf{90.23} & +0.06 \\
    Restaurant & 32.8 & 82.59 & \textbf{83.84} & +1.25 & 82.35 & 81.73 & -0.62 \\
    MultiNERD & 58.1 & 90.26 & \textbf{94.35} & +4.09 & 93.73 & 94.30 & +0.57 \\
    ncbi & 42.1 & 86.21 & 87.27 & +1.06 & 86.96 & \textbf{89.27} & +2.31 \\
    Ontonotes & 29.7 & 88.64 & \textbf{91.83} & +3.19 & 89.91 & 90.69 & +0.78 \\
    PolyglotNER & 33.6 & 53.31 & 66.90 & +13.59 & 65.67 & \textbf{67.52} & +1.85 \\
    TweetNER7 & 40.1 & 65.95 & \textbf{67.97} & +2.02 & 65.77 & 66.87 & +1.10 \\
    WikiANN & 52.0 & 64.47 & 85.19 & +20.72 & 84.91 & \textbf{86.87} & +1.96 \\
    wikiNeural & 57.7 & 88.27 & \textbf{93.71} & +5.44 & 93.28 & \textbf{93.71} & +0.43 \\
    \midrule
    \bf Avg & 34.9 & 81.53 & \textbf{86.15} & +4.62 & 85.07 & 86.09 & +1.02 \\
    \bottomrule
    \end{tabular}
    \caption{Supervised evaluation results. $\Delta$ indicates the improvement over the corresponding baseline. Results for InstructUIE and UniNER are derived from \citet{wang2023instructuie} and \citet{zhou2023universalner}, respectively.}
    \label{tab:supervised_evaluation}
\end{table*}

\begin{table}[!t]
    \centering
    \tabcolsep=1.5mm
    \footnotesize
    \begin{tabular}{lrccc}
    \toprule
    \bf Model & \bf \#Params. & \bf 0-shot & \bf Sup. & \bf Instance/s \\
    \midrule
    InstructUIE & 11B & 47.8 & 81.53 & 3.4 \\
    UniNER-7B & 7B & 53.4 & 85.07 & 1.6 \\
    \midrule
    GNER-T5-small & 77M & 48.2 & 77.43 & 32.5 \\
    GNER-T5-base & 248M & 59.5 & 83.21 & 20.2 \\
    GNER-T5-large & 783M & 63.5 & 85.45 & 11.5 \\
    GNER-T5-xl & 3B & 66.1 & 85.94 & 4.6 \\
    GNER-T5-xxl & 11B & 69.1 & 86.15 & 3.0 \\
    GNER-LLaMA & 7B & 66.1 & 86.09 & 4.0 \\
    \bottomrule
    \end{tabular}
    \caption{Model's performance and inference speed in zero-shot and supervised settings. The inference speed is tested in a single A100 node with batch size 4 per device. More details are outlined in Appendix~\ref{sec:detailed_results}.}
    \label{tab:supervised_evaluation2}
\end{table}

\subsection{Zero-shot Evaluation}

We evaluate the zero-shot performance of our models after the domain adaptation phrase.
Table \ref{tab:zero_shot_exp} summarizes the results.
Our model demonstrates significant improvements compared to other models. 
Significantly, although our GNER-LLaMA model shares the same backbone model (LLaMA-7B) and dataset (Pile-NER) with UniNER~\cite{zhou2023universalner}, it demonstrates a notable improvement.
Our results show that our 7B model outperforms the UniNER model of the same scale by approximately 12.7 $F_1$ score points on average, and exhibits improvements across every dataset.
Remarkably, our 7B model surpasses the UniNER 13B model by 10.5 points.
When considering smaller backbone models such as GNER-T5-base and GNER-T5-large, it's noteworthy that they also outperform all the aforementioned strong baselines.

\subsection{Supervised Evaluation}

To test our model's performance on supervised data, we conduct supervised multi-task fine-tuning based on the NER-specialized model.
The results are summarized in Table~\ref{tab:supervised_evaluation} and~\ref{tab:supervised_evaluation2}.
We first compare our approach with two closely related baselines, InstructUIE and UniNER, as we share the same backbone model and train with similar data. 
As a result, our method demonstrates significant improvements over these baselines: GNER-T5 achieves a 4.6-point increase in the $F_1$ score, while GNER-LLaMA shows a 1-point $F_1$ score improvement.
Moreover, we observe consistent enhancements across almost all datasets.
We also experiment with smaller models, considering both effectiveness and inference efficiency.
As shown in Table~\ref{tab:supervised_evaluation2}, our GNER-T5-large model, with only \textbf{10\%} the parameter size of UniNER, achieves superior performance and boasts \textbf{10$\times$} the inference efficiency.

\begin{table}
    \centering\small
    \begin{tabular}{l|r|r|r}
    \toprule
    Sequence Length & 0-60 & 60-100 & 100-200 \\
    \midrule
    LCS $O(N^2)$ & 1.0$\times$ & 1.0$\times$ & 1.0$\times$ \\
    LCS $O(N\log N)$ & 2.9$\times$ & 5.5$\times$ & 6.1$\times$ \\
    LCS (ours) & 3.8$\times$ & 12.6$\times$ & 17.3$\times$ \\
    \bottomrule
    \end{tabular}
    \caption{Acceleration effect of our optimized algorithm across different length ranges. All samples are selected from the generated results under supervised settings.}
    \label{tab:ablation_results1}
\end{table}

\subsection{Ablation Results}

We have demonstrated the effectiveness of negative instances in section~\ref{sec:preliminary}.
In this part, we conduct the ablation study to evaluate the performance of our LCS Matching Algorithm.
Our focus lies in two aspects: (1) how optimization in the algorithm increases the model's inference efficiency, and (2) how optimization in the matching condition improves the model's performance.
For the former, we evaluate the algorithm's acceleration across various sentence length ranges, as shown in Table~\ref{tab:ablation_results1}.
As sentence length increases, the acceleration effect of our algorithm becomes more pronounced. 
For sentence lengths between 100 and 200, it achieves an average acceleration factor of 17.3.
For the latter, we remove the Back Tokenization procedure from LCS and eliminate all LCS processes. The results, presented in Table~\ref{tab:ablation_results2}, indicate that removing Back Tokenization and the whole LCS algorithm leads to a decrease in effectiveness, underscoring the efficacy of our LCS Matching algorithm.

\begin{table}
    \centering\small
    \begin{tabular}{l|cc|cc}
    \toprule
    \multirow{2}{*}{Method} & \multicolumn{2}{c|}{GNER-T5-large} & \multicolumn{2}{c}{GNER-LLaMA} \\
     & 0-shot & Sup. & 0-shot & Sup. \\
    \midrule
    Ours & 63.47 & 85.45 & 66.07 & 86.09 \\
    w/o BT & 63.16 & 85.09 & 66.07 & 86.09 \\
    w/o LCS+BT & 62.31 & 84.91 & 65.77 & 85.99 \\
    \bottomrule
    \end{tabular}
    \caption{Ablation study of LCS Matching and Back Tokenization.}
    \label{tab:ablation_results2}
\end{table}
\section{Analysis}
\label{sec:analysis}

\begin{table}
    \small
    \centering
    \begin{tabular}{l|cccc}
        \toprule
        \bf Beam size & 1 & 2 & 3 & 4 \\
        \midrule
        UniNER-7B & 53.46 & 52.87 & - & - \\
        \midrule
        GNER-T5-base & 59.46 & 60.32 & 60.40 & 60.44 \\
        GNER-T5-large & 63.47 & 64.13 & 64.27 & 64.31 \\
        GNER-T5-xl & 66.12 & 66.81 & 66.86 & 66.88 \\
        GNER-T5-xxl & 69.06 & 69.20 & 69.33 & 69.33 \\
        LLaMA-7B & 66.07 & 66.87 & 67.00 & 67.08 \\
        \bottomrule
    \end{tabular}
    \caption{Zero-shot performance of UniNER and our model GNER via beam search.}
    \label{tab:beam_search_results}
\end{table}

\begin{figure}[!t]
    \begin{AcademicBox}[\footnotesize A Self-correction Example with beam size 2]
    \small
    \textbf{Token inputs:} What was the fog rated ? \\
    \textbf{Ground Truth:} \\
    What(O) was(O) the(B-title) fog(I-title) rated(O) ?(O) \\
    \vspace{-5pt} \hrule \vspace{4pt}
    \textbf{Medium prediction results} \\
    \textbf{highest beam score}: \\
    What(O) was(O) the(\textcolor{ErrorRed}{O}) fog(\textcolor{ErrorRed}{O}) \\
    \textbf{second-highest beam score}: \\
    What(O) was(O) the(\textcolor{CorrectGreen}{B-title}) fog(\textcolor{CorrectGreen}{I-title}) \\
    \vspace{-5pt} \hrule \vspace{4pt}
    \textbf{Final prediction results:} \\
    What(O) was(O) the(\textcolor{CorrectGreen}{B-title}) fog(\textcolor{CorrectGreen}{I-title}) rated(O) ?(O)
    \end{AcademicBox}
    \caption{An example of the self-correction mechanism when using beam search.}
    \label{fig:self_correction_example}
\end{figure}

\begin{figure}[!t]
    \centering
    \includegraphics[width=\linewidth]{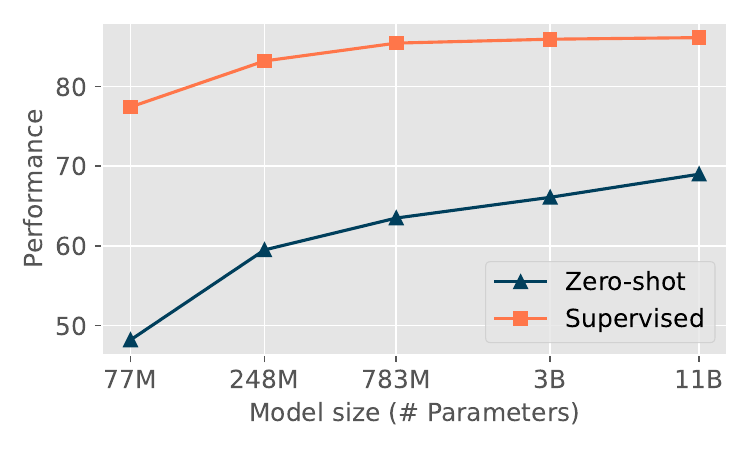}
    \caption{Scaling behavior of zero-shot and supervised performance with respect to model size (\# parameters).}
    \label{fig:scaling_law}
\end{figure}

\paragraph{Scaling Law of Generative NER Models}
Our experiments show that even smaller models like Flan-T5-large possess significant potential. 
We investigate the scaling law of Generative NER tasks in both zero-shot and supervised settings. The results are illustrated in Fig.~\ref{fig:scaling_law}. 
In the zero-shot setting, our methods scale well with model size. 
As the model size increases, the zero-shot capability of the model continues to rise, showing ample potential for further improvement with even larger models. 
In the supervised setting, our 783M model already demonstrates strong multi-task generalization abilities, and as the model size increases further, the improvements tend to converge.

\paragraph{Self-Correction Mechanism via Beam Search}
Beam search can enhance the performance of generative models by expanding the search space to include multiple hypotheses at each generation step.
Previous research~\cite{yan2021unified} has demonstrated that applying beam search in an entity-centric generation does not improve the model's performance or even degrade it.
We conduct experiments on UniNER and our model, the results of which are shown in Table~\ref{tab:beam_search_results}. 
We discover that as the beam size increases, the performance of the UniNER model decreases. 
In contrast, we observe a consistent improvement with beam search under our task schema. 
Upon a detailed case study of the model's generated results, we found that our task schema possesses a self-correction mechanism. 
The model retains some other hypotheses while generating subsequent results. 
In the decoding process that follows, the model can correct earlier mistakes.
As demonstrated in Fig.~\ref{fig:self_correction_example}, the model revises its previous incorrect prediction of ``O'' for ``the fog'' upon encountering the subsequent token ``rated''. 
This token is crucial for identifying the entity type associated with ``the fog''.
A detailed analysis is provided in Appendix~\ref{sec:self_correction_mechanism}.
\section{Conclusion}

This paper explores the potential of a strong Generative Named Entity Recognition (NER) system based on pre-trained LLMs by integrating the negative instances into training.
Through experiments, we have demonstrated significant advancements. 
Our approach, which combines the inclusion of contextual information and a clear definition of entity boundaries through negative instances, has proven to be highly effective in improving the model's performance, especially in zero-shot scenarios where prediction uncertainty is high.
The introduction of an LCS Matching algorithm further addresses the challenges of converting unstructured text into structured entities, ensuring accurate categorization and alignment. 
These findings highlight the crucial role of negative instances in NER tasks and the potential of generative models to revolutionize the field. 

\section{Limitation}

Despite our system achieving impressive results, there remain limitations and space for improvement.
In task settings, our approach focuses on the main-stream Flat-NER settings, where entities appear as continuous text segments, without addressing the discontinuous forms, i.e., discontinuous NER.
Actually, it has always been challenging for generative models to adopt a unified paradigm to resolve all the complex settings.
Previous entity-centric methods can address the discontinuous settings but fail to manage polysemy, where a phrase corresponds to different entity types in different sentence parts.
The primary focus of this paper is to explore the impact of negative instances in the training process, and we will explore a unified framework for generative models in future work.

\section*{Ethics Statement}
In this paper, we utilize the pre-trained large language models, i.e., Flan-T5 and LLaMA, as the foundational models. 
It's important to acknowledge that these models may contain inherent biases resulting from their pre-training processes. 
However, this issue is mitigated through our fine-tuning process, which refines the models to specifically concentrate on the Named Entity Recognition (NER) task, thereby reducing potential biases. 
Moreover, we strictly use all datasets and corpora in our study for scientific research purposes only. 

\section*{Acknowledgements}

We want to thank all the anonymous reviewers for their valuable comments. This work was supported by the National Science Foundation of China (NSFC No. 62206194), the Natural Science Foundation of Jiangsu Province, China (Grant No. BK20220488), Young Elite Scientists Sponsorship Program by CAST (2023QNRC001), and Supercomputing Center in Yancheng, Grant No. 20231001.

\bibliography{reference}
\bibliographystyle{acl_natbib}

\appendix

\clearpage
\section{Data Statistics and Pre-processing}
\label{sec:appendix_data}
We show the full dataset statistics in Table~\ref{tab:full_data_statistics}, including the domain of datasets, the number of instances in train/valid/test data, and their download address.
In particular, we have to pre-process the Pile-NER~\cite{zhou2023universalner} dataset to fit in our task schema.
We observe nuances between our compiled datasets and those referenced by \citet{wang2023instructuie} and \citet{zhou2023universalner}. 
Specifically, their MultiNERD and PolyglotNER datasets omit the last 10,000 training samples. 
Furthermore, they miss the last sample in some datasets for the validation and test sets, such as CrossNER politics, MIT Movie, and MIT Restaurant. 
We have included these omitted instances in our dataset, adhering to the original dataset compositions.
The modifications have a negligible impact on our results. 
This is because our sampling approach aligns with those used in the referenced studies, ensuring that the number of data instances sampled from each training set, up to 10,000 samples, is consistent.
Moreover, adding a single extra sample in the test sets hardly affects the final results.

\begin{table*}
\centering
\small
\begin{tabular}{l|c|c|ccc|c}
\toprule
\bf Dataset & \bf Domain & \bf Types & \bf \#Train & \bf \#Valid & \bf \#Test & \bf Download Link \\
\midrule
Pile-NER~\cite{zhou2023universalner} & \multirow{12}{*}{General} & 13,020 & 45,889 & 0 & 0 & \href{https://huggingface.co/datasets/Universal-NER/Pile-NER-type}{link} \\
CoNLL2003~\cite{sang2003introduction} & & 4 & 14,041 & 3,250 & 3,453 & \href{https://huggingface.co/datasets/conll2003}{link}\\
conllpp~\cite{wang2019crossweigh} & & 4 & 14,041 & 3,250 & 3,453 & \href{https://huggingface.co/datasets/conllpp}{link} \\
CrossNER AI~\cite{liu2021crossner} & & 13 & 100 & 350 & 431 & \href{https://github.com/zliucr/CrossNER/tree/main/ner_data/ai}{link} \\
CrossNER literature~\cite{liu2021crossner} & & 11 & 100 & 400 & 416 & \href{https://github.com/zliucr/CrossNER/tree/main/ner_data/literature}{link} \\
CrossNER music~\cite{liu2021crossner} & & 12 & 100 & 380 & 465 & \href{https://github.com/zliucr/CrossNER/tree/main/ner_data/music}{link} \\
CrossNER politics~\cite{liu2021crossner} & & 8 & 200 & 541 & 651 & \href{https://github.com/zliucr/CrossNER/tree/main/ner_data/politics}{link} \\
CrossNER science~\cite{liu2021crossner} & & 16 & 200 & 450 & 543 & \href{https://github.com/zliucr/CrossNER/tree/main/ner_data/science}{link} \\
MultiNERD~\cite{tedeschi-navigli-2022-multinerd} & & 16 & 144,144 & 10,000 & 10,000 & \href{https://huggingface.co/datasets/tner/multinerd}{link} \\
Ontonotes~\cite{weischedel2013ontonotes} & & 18 & 59,924 & 8,528 & 8,262 & \href{https://huggingface.co/datasets/tner/ontonotes5}{link} \\
PolyglotNER~\cite{al2015polyglot} & & 3 & 403,982 & 10,000 & 10,000 & \href{https://huggingface.co/datasets/polyglot_ner}{link} \\
WikiANN en~\cite{pan-etal-2017-cross} & & 3 & 20,000 & 10,000 & 10,000 & \href{https://github.com/Babelscape/wikineural/tree/master/data/wikiann/en}{link} \\
WikiNeural~\cite{tedeschi-etal-2021-wikineural-combined} & & 3 & 92,720 & 11,590 & 11,597 & \href{https://github.com/Babelscape/wikineural/tree/master/data/wikineural/en}{link} \\
\midrule
AnatEM~\cite{pyysalo2014anatomical} & \multirow{5}{*}{Biomed} & 1 & 5,861 & 2,118 & 3,830 & \href{http://nactem.ac.uk/anatomytagger/#AnatEM}{link} \\
bc2gm~\cite{smith2008overview} & & 1 & 12,500 & 2,500 & 5,000 & \href{https://github.com/spyysalo/bc2gm-corpus/tree/master/conll}{link} \\
bc4chemd~\cite{krallinger2015chemdner} & & 1 & 30,682 & 30,639 & 26,364 & \href{https://biocreative.bioinformatics.udel.edu/resources/biocreative-iv/chemdner-corpus}{link} \\
bc5cdr~\cite{li2016biocreative} & & 2 & 4,560 & 4,581 & 4,797 & \href{https://huggingface.co/datasets/ghadeermobasher/BC5CDR-Chemical-Disease}{link} \\
ncbi~\cite{dougan2014ncbi} & & 1 & 5,432 & 923 & 940 & \href{https://huggingface.co/datasets/ncbi_disease}{link} \\
\midrule
HarveyNER~\cite{chen-etal-2022-crossroads} & \multirow{5}{*}{Social media} & 4 & 3,967 & 1,301 & 1,303 & \href{https://github.com/brickee/HarveyNER/tree/main/data/tweets}{link} \\
Broad Tweet Corpus~\cite{derczynski-etal-2016-broad} & & 3 & 6,338 & 1,001 & 2,001 & \href{https://huggingface.co/datasets/strombergnlp/broad_twitter_corpus}{link} \\
TweetNER7~\cite{ushio-etal-2022-tweet} & & 7 & 7,111 & 886 & 576 & \href{https://huggingface.co/datasets/tner/tweetner7}{link} \\
mit-movie~\cite{liu2013asgard} & & 12 & 9,775 & 2,443 & 2,443 & \href{https://groups.csail.mit.edu/sls/downloads/movie}{link} \\
mit-restaurant~\cite{liu2013asgard} & & 8 & 7,660 & 1,521 & 1,521 & \href{https://groups.csail.mit.edu/sls/downloads/restaurant}{link} \\
\midrule
FabNER~\cite{kumar2022fabner} & STEM & 12 & 9,435 & 2,183 & 2,064 & \href{https://huggingface.co/datasets/DFKI-SLT/fabner}{link} \\
\midrule
FindVehicle~\cite{guan2023findvehicle} & Transportation & 21 & 21,565 & 20,777 & 20,777 & \href{https://github.com/GuanRunwei/FindVehicle}{link} \\
\bottomrule
\end{tabular}
\caption{Statistics of datasets in our collected datasets.}
\label{tab:full_data_statistics}
\end{table*}

\section{Hyper-parameters settings}
\label{sec:train_settings}

\begin{table*}[!t]
    \centering
    \small
    \begin{tabular}{c|c|c|c|p{10cm}}
    \toprule
    \bf Model & \bf Issue & \bf Case & \bf Type & \bf Prediction \\
    \midrule
    \multirow{10}{*}{\makecell{GNER \\ LLaMA}} & \multirow{4}{*}{\makecell{Omission \\ (39\%)}} & \multirow{2}{*}{1} & raw & who directed \textbf{the film} the lorax \\
    & & & pred. & who directed the lorax \\
    \cmidrule{3-5}
    & & \multirow{2}{*}{2} & raw & any reasonably priced indian restaurants in \textbf{the} theater district \\
    & & & pred. & any reasonably priced indian restaurants in theater district \\
    \cmidrule{2-5}
    & \multirow{2}{*}{\makecell{Addition \\ (3\%)}} & \multirow{2}{*}{3} & raw & the conservative regionalist navarra suma finished first and \,\ldots \\
    & & & pred. & the conservative regionalist \textbf{\textcolor{ErrorRed}{regionalist}}  navarra suma finished first and \,\ldots \\
    \cmidrule{2-5}
    & \multirow{4}{*}{\makecell{Substitution \\ (58\%)}} & \multirow{2}{*}{4} & raw & which five star italian restaurants in manattan have the best reviews \\
    & & & pred. & which five star italian restaurants in \textbf{\textcolor{ErrorRed}{manhattan}} have the best reviews \\
    \cmidrule{3-5}
    & & \multirow{2}{*}{5} & raw & polyethylene terephthalate ( pet ) bottles are made from ethylene and p-xylene . \\
    & & & pred. & polyethylene terephthalate \textbf{\textcolor{ErrorRed}{( p e t)}} bottles are made from ethylene and p-xylene . \\
    \midrule
    \multirow{16}{*}{\makecell{GNER \\ T5}} & \multirow{8}{*}{\makecell{Omission \\ (23\%)}} & \multirow{4}{*}{6} & raw & \ldots\, whose debut album tol cormpt norz norz \textbf{norz} rock hard journalist wolf-rüdiger mühlmann considers a part of war metal 's roots . \\
    & & & pred. & \ldots\, whose debut album tol cormpt norz norz rock hard journalist wolf-rüdiger mühlmann considers a part of war metal 's roots . \\
    \cmidrule{3-5}
    & & \multirow{4}{*}{7} & raw & jennifer lien starred in this action film of the \textbf{the} last six years that received a really good rating \\
    & & & pred. & jennifer lien starred in this action film of the last six years that received a really good rating \\
    \cmidrule{2-5}
    & \multirow{4}{*}{\makecell{Addition \\ (2\%)}} & \multirow{2}{*}{8} & raw & \ldots\, performed by wet wet wet that remained at number 1 \,\ldots \\
    & & & pred. & \ldots\, performed by wet wet wet \textbf{\textcolor{ErrorRed}{wet}} that remained at number 1 \,\ldots \\
    \cmidrule{3-5}
    & & \multirow{2}{*}{9} & raw & \ldots\, liked by many people that starred william forsythe \\
    & & & pred. & \ldots\, liked by many people that starred william forsythe \textbf{\textcolor{ErrorRed}{the}} \\
    \cmidrule{2-5}
    & \multirow{4}{*}{\makecell{Substitution \\ (75\%)}} & \multirow{4}{*}{10} & raw & four more children followed : charlotte brontë , ( 1816-1855 ) , branwell brontë ( 1817-1848 ) , emily brontë , ( 1818-1848 ) and anne ( 1820-1849 ) . \\
    & & & pred. & four more children followed : charlotte \textbf{\textcolor{ErrorRed}{bront}} , ( 1816-1855 ) , branwell \textbf{\textcolor{ErrorRed}{bront}} ( 1817-1848 ) , emily \textbf{\textcolor{ErrorRed}{bront}} , ( 1818-1848 ) and anne ( 1820-1849 ) . \\
    \bottomrule
    \end{tabular}
    \caption{Representative examples concerning the word addition, omission, and substitution problems in the zero-shot evaluation. We remove the label information in the predictions for a clear comparison with the raw texts.}
    \label{tab:long_sequence_problems}
\end{table*}

In our experiments, we train all models using a batch size of 256, employing the AdamW optimizer~\cite{loshchilov2018decoupled} for optimization.
For the T5 model, we set a constant learning rate of $5\times 10^{-5}$ and impose a length limitation of 640 tokens for both the encoder and decoder. For the LLaMA model, we adopt a cosine learning rate schedule, initiating with a warm-up phase that covers 4\% of the training steps, ramping up to a learning rate of $2\times 10^{-5}$, followed by a decay phase for the remainder of the training steps. 
The length limitation is set to 1280.
Due to our prediction sequences being longer, more training steps are required. The number of training epochs for our models varies by size: 20 epochs for both the small and base models, 10 epochs for the large and xl models, and 6 epochs for the xxl model. For the LLaMA model, we set the number of epochs to 3.
We observe an interesting phenomenon that the T5 model often requires more training steps to converge. 
A possible explanation is that the backbone model, Flan-T5, an instruction-tuned model without any Named Entity Recognition (NER) related data in the instruction-tuning process, requires more training steps to adapt to the NER task.

\section{Problems in long sequence}
\label{sec:longsequencepeoblem}
In response to the issues in long predictions mentioned in section~\ref{sec:lcs}, we conduct a detailed case study.
The representative examples are presented in Table~\ref{tab:long_sequence_problems}.
The problems can primarily be categorized into word omission, addition, and substitution, with omission and substitution accounting for the majority. 
We can conclude the following causes:

\paragraph{Noise in the original text}
Some of the issues can be attributed to noise in the original text. 
For example, in case 4, the model corrects ``manattan'' to ``manhattan'', and in case 7, it corrects the misuse of ``the''. 
However, we also observe that the model can introduce errors, as seen in cases 6 and 8, where the entities with repeated words, ``norz norz norz'' and ``wet wet wet'', confuse the model.

\paragraph{Missing words in the vocabulary}
Furthermore, we find that a certain proportion of issues can be derived from missing words in the model's vocabulary.
As a result, these words naturally do not appear in the model's output. 
For instance, in case 10, ``brontë'' was replaced with ``bront'' because ``brontë'' does not exist in the vocabulary. 
We also discovered that several special characters do not exist in the T5 vocabulary, leading to more occurrences of omission and substitution.

\paragraph{Accumulative exposure bias}
The issue of repetitive generation of words and phrases is common in long text generation (LTG) due to the accumulative exposure bias as the prediction length increases. 
As illustrated by cases 3 and 9, the model produces meaningless and repetitive information.

\section{Optimization in LCS Matching}
\label{sec:lcsmatching}
\paragraph{Optimization in Matching Algorithm}
We optimize the complexity of the LCS algorithm using a hierarchical divide-and-conquer approach through the following steps:
(1) If the sequence does not have the above problem, i.e., $\tilde{X} = X$, it is obvious that $\hat{Y} = \tilde{Y}$. The time complexity is $O(N)$, 
(2) For the omission case, where $\tilde{X}$ is a subsequence of $X$, the matching process can be accomplished in $O(N)$ through greedy matching.
(3) In other cases, we have implemented a fast version of the LCS algorithm~\cite{hunt1977fast} within $O(N\log N)$, based on the nature of the small number of duplicate words in $\tilde{X}$.
Our experimental results in Table~\ref{tab:ablation_results2} demonstrate that the optimization can significantly enhance efficiency, achieving up to a 17.3 times speedup for long sequences compared to the naive $O(N^2)$ implementation.

\paragraph{Back Tokenization}
One notable problem in the matching process is the missing words in the vocabulary, as detailed in our case study in Appendix~\ref{sec:longsequencepeoblem}.
For example, ``antropología'' in the original text becomes `antropologa' in the model's predictions, resulting in an inaccurate match in the matching process.
To address this, we employ back tokenization, which involves tokenizing each word in the original text and then detokenizing it to match a word in the model's vocabulary, thereby creating a more resilient matching condition.


\section{Self-correction Mechanism}
\label{sec:self_correction_mechanism}
In this section, we conduct a case study to explore (1) the reasons behind the reduced effectiveness of entity-centric methods like UniNER~\cite{zhou2023universalner} when beam search is applied, and (2) the specific enhancements of the self-correction mechanism in our task schema.
Upon comparing UniNER's performance with and without beam search, we observe that beam search leads to the model responding with the same answers across a variety of entity-type queries.
For our models, we provide representative examples in Table~\ref{tab:self_correction_examples} to illustrate the self-correction mechanism's impact, showcasing (1) enhanced precision in determining entity boundaries (cases 1 and 2), (2) the use of contextual clues to recognize inherent entities (cases 3, 6 and 7), and correct mistakes (cases 4 and 5).

\section{Detailed Evaluation Results}
\label{sec:detailed_results}
We detail the performance of our models across all datasets in Table~\ref{tab:appexdix_detailed_results}, including error bars for zero-shot performance derived from the variance of five separate runs.
For the supervised settings, we do not conduct multiple runs due to the extensive size of the datasets, where the training and inference process can be very time-consuming.
Our trials with smaller models indicate that the variability, or error bars, for models in the supervised settings is minimal, approximately around 0.15.

\begin{table*}[!t]
    \centering
    \small
    \begin{tabular}{c|c|c|p{10.5cm}}
    \toprule
    \bf Model & \bf Case & \bf Type & \bf Text Generations \\
    \midrule
    \multirow{12}{*}{\makecell{GNER \\ LLaMA}} & \multirow{2}{*}{1} & w/o beam search & who(O) is(O) directing(O) \textcolor{ErrorRed}{\bf the(O)} hobbit(B-title) \\
    \cmidrule{3-4}
    & & w/ beam search & who(O) is(O) directing(O) \textcolor{CorrectGreen}{\bf the(B-title) hobbit(I-title)} \\
    \cmidrule{2-4}
    & \multirow{2}{*}{2} & w/o beam search & what(O) is(O) the(O) plot(O) of(O) \textcolor{ErrorRed}{\bf the(O)} wild(B-title) bunch(I-title) \\
    \cmidrule{3-4}
    & & w/ beam search & what(O) is(O) the(O) plot(O) of(O) \textcolor{CorrectGreen}{\bf the(B-title) wild(I-title) bunch(I-title)} \\
    \cmidrule{2-4}
    & \multirow{2}{*}{3} & w/o beam search & was(O) there(O) a(O) \textcolor{ErrorRed}{\bf romantic(O) film(O) noir(O)} \\
    \cmidrule{3-4}
    & & w/ beam search & was(O) there(O) a(O) \textcolor{CorrectGreen}{\bf romantic(B-genre) film(I-genre) noir(I-genre)} \\
    \cmidrule{2-4}
    & \multirow{2}{*}{4} & w/o beam search & does(O) paymon(B-Restaurant Name) serves(O) \textcolor{ErrorRed}{\bf white(B-Cuisine) wine(I-Cuisine)} \\
    \cmidrule{3-4}
    & & w/ beam search & does(O) paymon(B-Restaurant Name) serves(O) \textcolor{CorrectGreen}{\bf white(B-Dish) wine(I-Dish)} \\
    \midrule
    \multirow{8}{*}{\makecell{GNER \\ T5}} & \multirow{2}{*}{5} & w/o beam search & \ldots\, some(O) \textcolor{ErrorRed}{\bf batman(B-character)} movies(O) from(O) the(O) 1990s(B-year) \\
    \cmidrule{3-4}
    & & w/ beam search & \ldots\, some(O) \textcolor{CorrectGreen}{\bf batman(B-title) movies(I-title)} from(O) the(O) 1990s(B-year) \\
    \cmidrule{2-4}
    & \multirow{2}{*}{6} & w/o beam search & where(O) was(O) \textcolor{ErrorRed}{\bf the(O)} presidio(B-title) filmed(O) \\
    \cmidrule{3-4}
    & & w/ beam search & where(O) was(O) \textcolor{CorrectGreen}{\bf the(B-title) presidio(I-title)} filmed(O) \\
    \cmidrule{2-4}
    & \multirow{2}{*}{7} & w/o beam search & \ldots\, the(O) \textcolor{ErrorRed}{\bf third(O) harry(O) potter(O) movie(O)} called(O) \\
    \cmidrule{3-4}
    & & w/ beam search & \ldots\, the(O) \textcolor{CorrectGreen}{\bf third(B-title) harry(I-title) potter(I-title) movie(I-title)} called(O) \\
    \midrule
    \bottomrule
    \end{tabular}
    \caption{Representative examples in the self-correction mechanism via beam search.}
    \label{tab:self_correction_examples}
\end{table*}

\section{Environmental Impact}
Training huge models can have a negative impact on the environment.
All our models are trained on the hardware of a single A100 node (8$\times$ Nvidia-A100-80G-SXM4) with approximately 800 GPU hours in total.
The carbon footprint estimation is 135.3 kg $\text{CO}_2$eq according to \citet{wu2022sustainable}.

\begin{table*}[!t]
    \centering
    \footnotesize
    \begin{tabular}{l|cccccc}
    \toprule
    \bf Method & GNER-T5 & GNER-T5 & GNER-T5 & GNER-T5 & GNER-T5 & GNER-LLaMA \\
    \bf Backbone & Flan-T5-small & Flan-T5-base & Flan-T5-large & Flan-T5-xl & Flan-T5-xxl & LLaMA-7B \\
    \bf \# Params. & 77M & 248M & 783M & 3B & 11B & 7B \\
    \midrule
    \multicolumn{7}{c}{\bf Zero-shot Performance} \\
    \midrule
    AI & 50.18$\pm$0.9 & 56.83$\pm$0.4 & 62.56$\pm$0.2 & 62.09$\pm$0.3 & 68.19$\pm$0.3 & 63.11$\pm$0.2 \\
    Literature & 49.78$\pm$1.5 & 58.68$\pm$0.8 & 58.20$\pm$0.4 & 64.94$\pm$1.1 & 68.66$\pm$0.2 & 68.20$\pm$0.3 \\
    Music & 65.83$\pm$1.3 & 72.29$\pm$0.3 & 76.73$\pm$0.7 & 80.59$\pm$0.6 & 81.24$\pm$0.4 & 75.72$\pm$0.8 \\
    Politics & 57.28$\pm$1.1 & 64.50$\pm$1.1 & 66.99$\pm$0.8 & 73.73$\pm$0.6 & 75.11$\pm$0.9 & 69.38$\pm$1.2 \\
    Science & 62.68$\pm$1.9 & 68.00$\pm$1.2 & 72.60$\pm$0.2 & 68.74$\pm$1.2 & 76.70$\pm$1.0 & 69.93$\pm$0.4 \\
    Movie & 37.38$\pm$1.8 & 54.52$\pm$0.2 & 58.59$\pm$0.1 & 62.96$\pm$0.4 & 62.52$\pm$0.5 & 68.63$\pm$0.5 \\
    Restaurant & 14.30$\pm$1.4 & 41.41$\pm$1.2 & 48.61$\pm$0.5 & 49.82$\pm$0.2 & 51.04$\pm$0.4 & 47.49$\pm$1.1 \\
    \midrule
    \bf Avg. & 48.20$\pm$1.1 & 59.46$\pm$0.8 & 63.47$\pm$0.2 & 66.12$\pm$0.2 & 69.06$\pm$0.3 & 66.07$\pm$0.3 \\
    \midrule
    \multicolumn{7}{c}{\bf Supervised Performance} \\
    \midrule
    AnatEM & 81.02 & 86.99 & 90.22 & 90.29 & 90.30 & 90.24 \\
    bc2gm & 69.02 & 79.11 & 83.10 & 84.25 & 84.29 & 83.18 \\
    bc4chemd & 76.33 & 85.19 & 88.51 & 90.22 & 90.04 & 89.40 \\
    bc5cdr & 82.02 & 87.16 & 88.81 & 89.83 & 89.95 & 90.27 \\
    Broad Twitter & 80.09 & 81.59 & 82.61 & 84.34 & 84.56 & 83.74 \\
    CoNLL2003 & 89.12 & 91.82 & 93.14 & 93.14 & 93.28 & 93.60 \\
    FabNER & 68.20 & 77.34 & 81.89 & 81.54 & 83.20 & 85.39 \\
    FindVehicle & 90.64 & 93.61 & 95.71 & 95.97 & 97.37 & 98.62 \\
    HarveyNER & 60.27 & 70.77 & 75.24 & 74.00 & 76.33 & 74.73 \\
    Movie & 85.03 & 88.08 & 89.39 & 89.31 & 89.28 & 90.23 \\
    Restaurant & 78.98 & 82.21 & 83.72 & 83.06 & 83.84 & 81.73 \\
    MultiNERD & 90.94 & 93.17 & 94.24 & 94.51 & 94.35 & 94.30 \\
    ncbi & 82.06 & 87.14 & 88.46 & 89.58 & 88.27 & 88.55 \\
    Ontonotes & 86.36 & 89.33 & 90.54 & 91.63 & 91.83 & 90.69 \\
    PolyglotNER & 45.27 & 62.13 & 66.16 & 67.15 & 66.90 & 67.52 \\
    TweetNER7 & 62.92 & 67.36 & 67.50 & 68.07 & 67.97 & 66.87 \\
    WikiANN & 76.58 & 82.56 & 85.32 & 86.09 & 85.19 & 86.87 \\
    wikiNeural & 88.97 & 92.24 & 93.56 & 93.85 & 93.71 & 93.71 \\
    \midrule
    \bf Avg. & 77.43 & 83.21 & 85.45 & 85.94 & 86.15 & 86.09 \\
    \bottomrule
    \end{tabular}
    \caption{Zero-shot and supervised evaluation results.}
    \label{tab:appexdix_detailed_results}
\end{table*}

\end{document}